\documentclass{article}

\usepackage{arxiv}
\usepackage{amsthm}
\usepackage[utf8]{inputenc} 
\usepackage[T1]{fontenc}    
\usepackage{algorithm} 
\usepackage{algpseudocode,color}
\usepackage{hyperref}       
\usepackage{url}            
\usepackage{booktabs}       
\usepackage{amsfonts}       
\usepackage{nicefrac}       
\usepackage{microtype}      
\usepackage{lipsum}		
\usepackage{graphicx}
\usepackage[square,sort,comma,numbers]{natbib}
\usepackage{doi}
\graphicspath{{./Arxiv/figs}}

\newtheorem{theorem}{Theorem}

\newtheorem{problem}{Problem}

\title{Multi-robot Searching with Limited Sensing Range for Static and Mobile Intruders}

\author{ \href{https://orcid.org/0000-0002-0245-0967}{\includegraphics[scale=0.06]{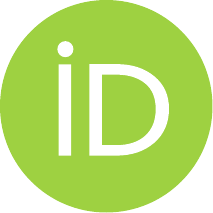}\hspace{1mm}Swadhin Agrawal}\thanks{Swadhin Agrawal is with IISER  Bhopal, Bhopal, India. Email: {\tt\small swadhin20@iiserb.ac.in}} \\
	Department of Electrical Engineering and Computer Science\\
	Indian Institute of Science Education and Research,\\
	Bhopal, India \\
	\texttt{swadhin20@iiserb.ac.in} \\
	\And
	\href{https://orcid.org/0000-0003-0104-1659}{\includegraphics[scale=0.06]{orcid.pdf}\hspace{1mm}Sujoy Bhore}\thanks{Sujoy Bhore is with IIT Bombay, Mumbai, India. Email: {\tt\small sujoy@cse.iitb.ac.in}} \\
	Department of Computer Science\\
	Indian Institute of Technology,\\
	Mumbai, India \\
	\texttt{sujoy@cse.iitb.ac.in} \\
	\And
    \href{https://orcid.org/0000-0002-0152-2279}{\includegraphics[scale=0.06]{orcid.pdf}\hspace{1mm}Joseph S.B. Mitchell}\thanks{Joseph S. B. Mitchell is with the Department of Applied Mathematics and Statistics, Stony Brook University, United States. Email: {\tt\small joseph.mitchell@ stonybrook.edu}} \\
	Department of Applied Mathematics and Statistics\\
	Stony Brook University,\\
	United States\\
	\texttt{joseph.mitchell@ stonybrook.edu} \\
	\And
 \href{https://orcid.org/0000-0002-7297-1493}{\includegraphics[scale=0.06]{orcid.pdf}\hspace{1mm}P.B. Sujit}\thanks{P.B. Sujit is with IISER  Bhopal, Bhopal, India. Email: {\tt\small sujit@iiserb.ac.in}} \\
	Department of Electrical Engineering and Computer Science\\
	Indian Institute of Science Education and Research,\\
	Bhopal, India \\
	\texttt{sujit@iiserb.ac.in} \\
	\And
  \href{}{Aayush Gohil}\thanks{Aayush Gohil is with Nirma University, Ahmedabad, India. Email: {\tt\small aayushgohil3433@gmail.com}} \\
	Department of Computer Science\\
	Nirma University,\\ Ahmedabad, India \\
	\texttt{aayushgohil3433@gmail.com} \\
}

\renewcommand{\shorttitle}{Multi-robot Searching with Limited Sensing Range for Static and Mobile Intruders}

\hypersetup{
pdftitle={Multi-robot Searching with Limited Sensing Range for Static and Mobile Intruders},
pdfsubject={q-bio.NC, q-bio.QM},
pdfauthor={Swadhin Agrawal, Sujoy Bhore, Joseph S. B. Mitchell, P.B. Sujit, Aayush Gohil},
pdfkeywords={First keyword, Second keyword, More},
}

\begin{document}
\maketitle

\begin{abstract}
We consider the problem of searching for an intruder in a geometric domain by utilizing multiple search robots. The domain is a simply connected orthogonal polygon with edges parallel to the cartesian coordinate axes. Each robot has a limited sensing capability. 
We study the problem for both static and mobile intruders. It turns out that the problem of finding an intruder is \textsf{NP}-hard, even for a stationary intruder. Given this intractability, we turn our attention towards developing efficient and robust algorithms, namely methods based on space-filling curves, random search, and cooperative random search. Moreover, for 
each proposed algorithm, 
we evaluate the trade-off between the number of search robots and the time required for the robots to complete the search process while considering the geometric properties of the connected orthogonal search area. 
\end{abstract}


\section{Introduction}
A central problem for the surveillance and monitoring of a domain is that of devising efficient strategies for a team of searchers to intercept an intruder (a target) within the domain. The problem is well motivated by applications in securing sensitive installations, including military bases, nuclear power plants, border patrol, natural habitat conservation, etc. The use of robots is ubiquitous for these types of purposes. This class of problems has been investigated extensively in the literature over the decades~\cite{parsons2006pursuit,kolling2010,suzuki1992,sachs2004,LaValle2001,guibas1999visibility,park2001}. Moreover, many practical algorithms have been developed that are used in real-life scenarios~\cite{zhang2014defending,haas2018optimal}. 

With recent developments in sensing and electronic technology, the deployment of multi-robot systems has become a reality. However, the sensing capabilities are still limited. The limited sensing range raises critical concerns, particularly in the context of intruder interception. Therefore, it is evident that local geometric properties, such as the shape and structure of a region, can be used to calibrate the effectiveness of the search process. 

Motivated by this, we consider the following problem in this work: considering a closed search space, we model the region as a simply connected orthogonal polygon in $\mathbb{R}^2$. The objective is to deploy a set of $k$ robots with limited but uniform sensing capacities, such that if there is a stationary or mobile intruder inside the region, then the robots will be guaranteed to intercept the intruder as quickly as possible. Each of the $k$ robots can be deployed to an arbitrary location in the domain; this location serves as the starting point for a path that the robot will execute. Our goal is to compute these starting points and the corresponding paths so that the length of the longest path is minimized while assuring that the paths intercept the intruder within $\phi$-steps, thereby leading to the discovery of the intruder. Clearly, if one deploys a very large number of robots, enough so that the sensing regions of the robots at their deployment locations completely cover the domain, then any intruder is detected instantaneously. However, in a more realistic scenario, one has a limited number of robots, and therefore, the robots must be moved strategically in order to detect an intruder, so the objective is to minimize the ``makespan'' (maximum length of the robot paths) so that the search is concluded as soon as possible without having to search the entire domain (in most of the cases) while using only the $k$ robots that are available to the user. We seek to determine the trade-off between the number, $k$, of robots provided and the total time, $T$, it takes to execute the $k$ search paths in order to assure the interception of the intruder and determine how the trade-off changes with the considered geometric properties of the orthogonal search area. 

\section{Related Work}

The problem of placing a small number of static sensors within a geometric domain in order to achieve visibility coverage (in which a point is ``covered'' by the sensors if it is within line of sight of any sensor, regardless of the distance to the sensor) is commonly called an ``art gallery problem'' or a ``guarding problem''; this class of problems has been extensively studied, particularly in the field of computational geometry, e.g.,~\cite{urrutia2000art,abrahamsen2021art,lee1986computational,hoffmann1990rectilinear}. Most variants of this class of geometric set cover problems that are intractable (NP-hard and often also hard to approximate). The usual model assumes no limit on the range of sensing, a constraint we impose in our setting in this paper. 
In the case that there is a mobile ``guard'', a single guard is enough to cover a connected domain, and the \emph{watchman route problem}~\cite{chin1986optimum,delm-tsp-03,mitchell2013approximating} seeks to minimize the length of a route in order that all points of the domain are seen by the guard. (In the $k$-watchman route problem, one seeks to optimize the routes of $k$ mobile guards, either to minimize the longest of the routes or to minimize the sum of the route lengths.)

Another related problem that is well studied is the \emph{cops-robber} game on graphs, where a robber is stationed at different vertices of an input graph, and the objective is to determine the minimum number of cops that is sufficient to catch the robber. 
Both cops and robbers move around the graph, and the goal of the cops is to intercept the robber. 
The problem is \textsf{NP}-hard on general graphs, and there exists an approximation scheme~\cite{abraham2019cops}. However, it is not known if the problem remains hard on special graph families, e.g., 
grid graphs within a simply connected domain (see also~\cite{bollobas2013cops,bonato2011game,na2007cops,bonato2009capture}). In a continuous domain, another problem related to \emph{cops-robber} is the class of \emph{pursuit evasion} problems, including the visibility-based pursuit-evasion problem, for which bounds have been studied on the number of pursuers needed in order to ensure capture of a mobile intruder~\cite{guibas1999visibility,klein2012catch}. Similar work that focuses on mobile intruders (see~\cite{Hollinger2009}) provides scalable solutions. However, the performance of such solutions can not be guaranteed when the intruder's motion model is unknown or when the intruder is adversarial. This raises the need to develop robust multi-agent searching algorithms for worst-case and unknown target motion models.

\section{Problem Formulation}

\begin{figure}
    \centering
    \begin{minipage}{0.45\linewidth}
    \includegraphics[scale=0.6]{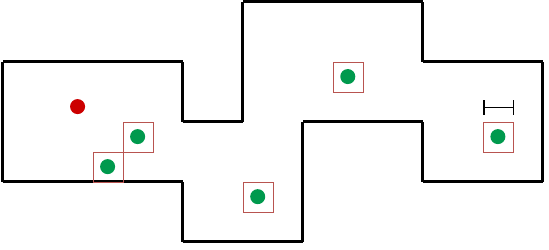}
    \put(-68,28){\footnotesize{\textbf{$\mathcal{P}$}}}
    \put(-147,45){\footnotesize{\textbf{$\mathcal{I}$}}}
    \put(-131,33){\footnotesize{\textbf{$r_1$}}}
    \put(-142,25){\footnotesize{\textbf{$r_2$}}}
    \put(-97,15){\footnotesize{\textbf{$r_3$}}}
    \put(-70,50){\footnotesize{\textbf{$r_4$}}}
    \put(-27,27){\footnotesize{\textbf{$r_5$}}}
    \put(-18,45){\footnotesize{\textbf{$l^s$}}}\hspace{0.5cm}
    \put(-90,-15){(a)}
    \hspace{0.5cm}
    \end{minipage}
    \begin{minipage}{0.45\linewidth}
    \includegraphics[scale=0.7]{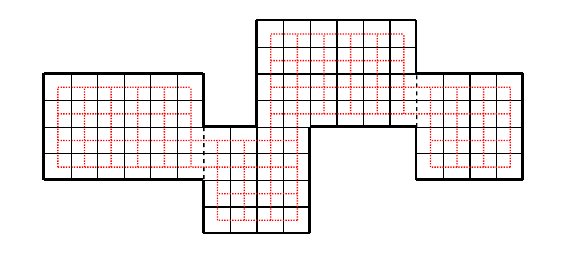}
    \put(-82,33){\footnotesize{\textbf{$G(\mathcal{P})$}}}
    \put(-92,-8){(b)}
    \end{minipage}
    \caption{(a) Polygonal region $\mathcal{P}$ with intruder ($\mathcal{I}$) in red color and the search robots ($\mathcal{R'}\subseteq \mathcal{R}$) in green. The red square boxes around the robots is the robot sensing range, $l^s$. (b) The grid graph $G(\mathcal{P})$ shown in black solid lines of the region. 
    The dual of $G(\mathcal{P})$, $\mathcal{D}_G$, is shown with red dotted lines. The robots move (horizontally/vertically) from the center of each black pixel (square) to the center of another at each time step $t$.} 
    \label{fig:illustration}
\end{figure}
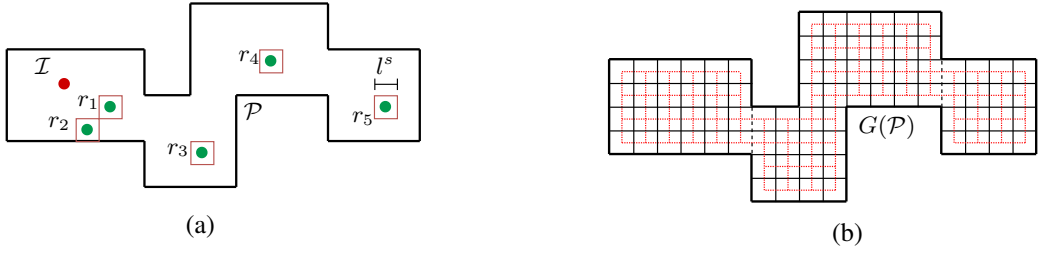

Consider a simply connected orthogonal polygon $\mathcal{P}$ as shown in Fig.~\ref{fig:illustration}(a) having integral coordinates. The polygon $\mathcal{P}$ can be considered to be the union of a finite number of unit square pixels (together considered as a grid-graph, $G(\mathcal{P})$) (black squares in Fig.~\ref{fig:illustration}(b)). Let $\mathcal{I}$ be an intruder (red dot in Fig.~\ref{fig:illustration}(a)) that is present in the region $\mathcal{P}$. Let $\mathcal{R}=\{r_1,\ldots,r_k\}$, for some $k\in \mathbb{N}$, be a 
set of robots (green dots in Fig.~\ref{fig:illustration}(a)), each of which is modelled as a point mass and deployed to search the intruder, $\mathcal{I}$ in $\mathcal{P}$. Each robot $r_i\in \mathcal{R}$ moves from one-pixel center to another, at each integral time step $t$ moving one unit vertically or horizontally (along the edges of the dual graph $\mathcal{D}_G$ (red dotted lines in Fig.~\ref{fig:illustration}(b)) of $G(\mathcal{P})$). We consider the sensing range of a robot to be 1/2 in the $L_\infty$ metric so that when a robot is at the center of a pixel, it senses the unit area of the square pixel, detecting and capturing the intruder if it is within the pixel as depicted in Fig.~\ref{fig:illustration}(a) with the red square region around the robots. Due to the limited sensing range in this formulation, multiple robots can be on the same pixel at any given instance of time. In what follows, we define the problems formally.

\begin{problem}
\textsc{Multi-Robot Search (MRS):}
Given the starting locations of $k$ robots in a simply connected orthogonal polygonal region $\mathcal{P}$ having $n$ vertices, the goal is to intercept an intruder $I$ in $\phi$ steps, where $I$ is stationed at any unknown point in $\mathcal{P}$.   
%
%
\end{problem}

\begin{problem}\textsc{Dynamic Multi-Robot Search (\textsc{DMRS}):}
Given the starting locations of $k$ robots in a simply connected orthogonal polygonal region $\mathcal{P}$ having $n$ vertices, the goal is to decide if it is possible to intercept an intruder $I$ in $\phi$ steps, where $I$ is 
initially stationed at any unknown point in $\mathcal{P}$ and is moving inside $\mathcal{P}$.   
%
\end{problem}

We assume that the intruder has perfect knowledge of where every robot is at every moment (the intruder has no limit on its sensing range). While the intruder is assumed to have positional knowledge of the robots, it is assumed not to have the power to learn the movement patterns of robots in a manner that would allow it to predict future locations and trajectories of robots.

Given the search scenario, it is NP-hard to determine an algorithm that can guarantee the solution to \textit{Problem 1} since the length of the paths can increase exponentially with respect to the number of cells. Further, in the case of \textit{Problem 2}, a similar argument can be extended by considering the emergence and disappearance of the cells based on the ones that are not yet searched and the ones that are previously searched (such searched cells can re-emerge as recontaminated cell as the time progresses), respectively. Thus, in the following section, we describe the NP-hardness nature of the problem formally by reducing the \textit{Problem 1} and \textit{Problem 2} to the previously known NP-hard, $3$-\textsc{Partition}, problem~\cite{Garey1978}.

\section{Hardness Results}\label{sec:hardness}

Let $\mathcal{P}$ be a simple orthogonal polygon, and $G_{\mathcal{P}}$ be the underlying grid graph on $\mathcal{P}$. We are given $k$ robots that move with the same constant speed, i.e., at each time step, each robot can move from the current vertex to any of its four neighbouring vertices.
There is an intruder $\mathcal{I}$ that is present on a vertex of $G_{\mathcal{P}}$. The objective is to discover $\mathcal{I}$ within a minimum number of steps. In what follows, we show that the \textsc{MRS} is \textsf{NP}-hard, by showing a reduction from the $3$-\textsc{Partition} problem. 

\begin{theorem}
\textsc{MRS} on a grid graph is \textsf{NP}-hard. 
\end{theorem}

\begin{figure}
     \centering
  
      \includegraphics[scale=0.4]{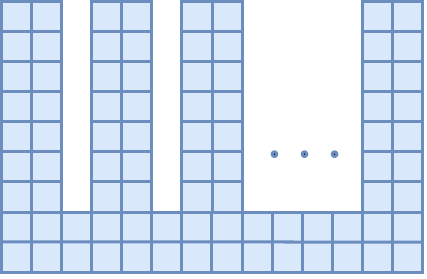}
    \put(-100,35){\footnotesize{\textbf{$\mathcal{P}$}}}
    \put(-66,5){\footnotesize{\textbf{$b$}}}
    \put(-80,60){\footnotesize{\textbf{$s_1$}}}
    \put(-63,60){\footnotesize{\textbf{$s_2$}}}
    \put(-45,60){\footnotesize{\textbf{$s_3$}}}
    \put(-12,60){\footnotesize{\textbf{$s_{3q}$}}} \caption{Comb structure for the polygon $\mathcal{P}$. The base $b$, of the structure has a diameter $O(q)$. The spikes $s_0,s_1, \ldots, s_{3q}$ represent sub-regions that are connected to the base. Each spike has a width that is equal to the shortest edge of the polygon that appears in that spike.}  \label{fig:nphardness}
\end{figure}
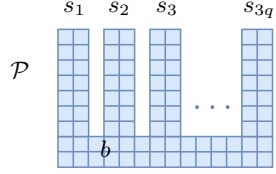

\begin{proof}
Let $\mathcal{F}$ be an instance of the $3$-\textsc{Partition} problem, where the input is a multiset of positive integers, $S$, with $|S| = 3q$.
The objective is to find if there exists a partition of $S$ into $q$ triplets $S_1, S_2, \ldots, S_q$ such that the sum of the numbers in each one is equal to $T$. Hence, the sum of all triplets is $qT$. Moreover, the $S_1, S_2, \ldots, S_q$ must form a disjoint partition of $S$, and their union result in the whole set $S$. This $3$-\textsc{Partition} problem is known to be strongly \textsf{NP}-Complete.

Now, consider an instance $\mathcal{W}$ of \textsc{MRS}. The orthogonal polygon $\mathcal{P}$ can be mapped onto a comb-like structure (see Fig.~\ref{fig:nphardness}) that consists of a base and $\beta = 3q$ number of spikes. The base $\textit{b}$ is a rectangle with the diameter $O(q)$. 
Let each spike $s_i$ be represented by a distinct number $n_i$ that corresponds to the time taken by $k$ robots to clear the $i^{th}$ spike. The objective is to determine the triplets ($n_i + n_j + n_k = T$) of such spikes so that the total time taken by $k$ robots to clear all the triplets is $qT$.


In the forward direction, we show that if there is a solution to the $3$-\textsc{Partition} problem, then we can obtain a solution for the \textsc{MRS} problem in polynomial time. Consider a triplet $S_i=\{n_x,n_y,n_z\}$ from the solution of $3$-\textsc{Partition}. We assign the $k$ robots to traverse the spikes $\{s_x,s_y,s_z\}$ corresponding  to $\{n_x,n_y,n_z\}$, respectively. The traversal follows the order of the spikes from left to right. For each triplet, we follow the same procedure. Since the sum of $\{n_x,n_y,n_z\}$ is $T$, it will take $qT$ time for a robot to clear the corresponding spikes. We shall obtain the solution to the MRS problem in $qT$ steps as searching is done simultaneously with traversal in the spikes. 

Conversely, let there exists a solution of \textsc{MRS} that can be obtained in polynomial time, which means that the robots can be assigned to exactly three spikes such that the sum of their search time is $qT$. We can consider the corresponding numbers as a triplet and report all such triplets as a solution to the $3$-\textsc{Partition} problem. 
This concludes the proof.
\end{proof}

\paragraph{Remark}
In the proof, we did not explicitly prohibit multiple 
robots from being on the same location at any particular time. However, it is easy to adjust the proof even for the case, when it is not allowed to have two robots at the same location at the same time. For each robot which is assigned to clear a spike (say, $s_i$), we enforce that the robot has to wait for $n_i$ amount of time.  

\paragraph{Remark}
We conjecture that the dynamic version (DMRS), with a mobile intruder, is also \textsf{NP}-hard, perhaps using a similar reduction.

\section{Algorithmic Methods}\label{sec:algorithms}

Given the hardness status of the problem, in this section, we propose three algorithmic approaches, namely, (i) Space-filling curves, (ii) Random search, and (iii) Cooperative random search, for robots to search a static/mobile intruder in $\mathcal{P}$.

\subsection{Space-filling-curves approach}\label{sec:sfc}

A space-filling curve $\mathcal{C}$ is a mathematical curve on grid graphs spanning a simply connected closed polygonal region without intersecting or repeating itself. It is not known how to compute a space-filling curve for any simply connected non-convex orthogonal polygonal region in $\mathbb{R}^2$ ~\cite{sagan2012space}. Hence, in this approach, we rectangulate the input polygon $\mathcal{P}$ to decompose it into disjoint rectangles $d_1, \ldots, d_{q}$, for some $q\in \mathbb{N}$. Subsequently, we compute the space-filling curve (whose edges form a subset of the edge set of the dual-graph, $\mathcal{D}_G$) for each of the rectangles by utilizing the grid graph of the rectangular sub-region. Then, we distribute robots proportionately to the area of the rectangles. Each robot is assigned to a pre-computed fraction (the robot moves from its start position until the start position of its neighbouring adjacent robot) of the space-filling curve. Finally, the robots simultaneously search in all the rectangles by moving along the assigned paths.

\noindent\paragraph{Rectangulation} A simple method for rectangulation is to shoot vertical and horizontal rays from each boundary point of an orthogonal polygon. However, this process can create a quadratic number of cells, which would require at least quadratic many robots to continue with a parallel search. Therefore, we use a different method for the rectangulation. 

\begin{figure}
    \centering
    \includegraphics[scale=0.3]{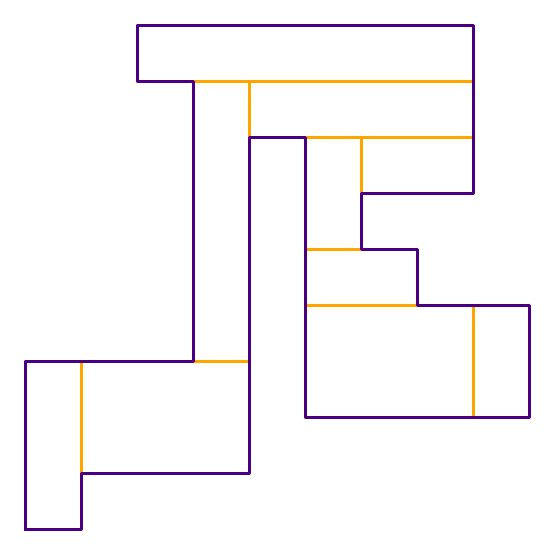}
    \includegraphics[scale=0.3]{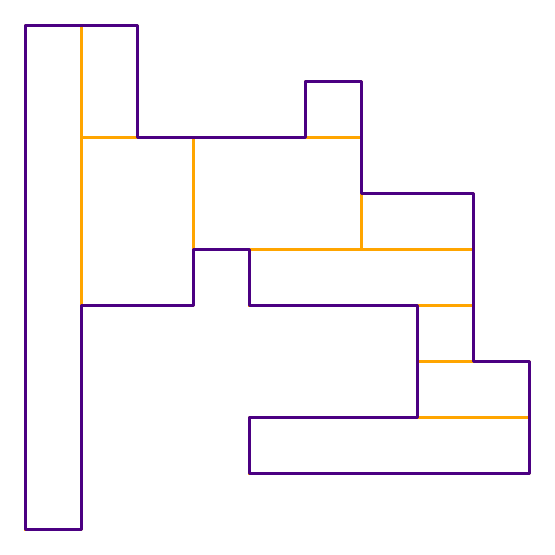}
    \put(-2.7cm,-0.5cm){(a)}
    \hspace{0.2cm}
    \raisebox{0.5cm}{\includegraphics[scale=0.3]{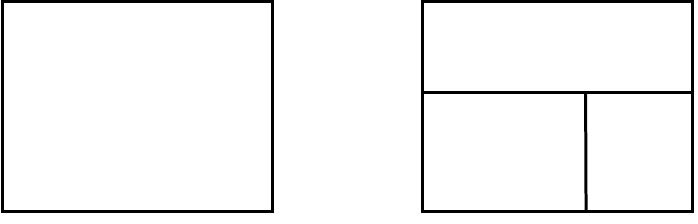}
    \put(-85,-6){$a$}
    \put(-107,8){$b$}
    \put(-47,5){$\frac{b}{2}$}
    \put(-35,-6){\footnotesize$a/2$}}
    \put(-2cm,-0.5cm){(b)}
    \caption{(a) Rectangular decomposition of simply connected rectilinear polygons, (b) Splitting of rectangle for computing space-filling curves}
    \label{fig:decomposition}
\end{figure}
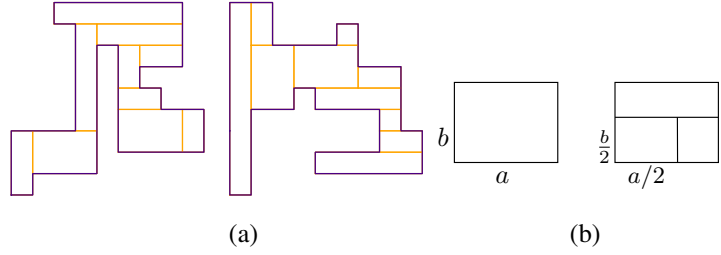

We assume that $\mathcal{P}$ is provided as a list of vertices along the boundary. Consider a bounding rectangle $\mathcal{B}$ that contains $\mathcal{P}$. We overlay a grid of side length $l^b$ on $\mathcal{B}$, where $l^b$ is the shortest edge of the boundary of $\mathcal{P}$. 
We build a grid graph, $\mathcal{T}$, by considering the grid cells that are inside the polygon $\mathcal{P}$. 
In order to select grids that lie inside $\mathcal{P}$, for each grid cell, we shoot vertical and horizontal rays in all four directions from the center of the cell. We count the number of times each line intersects the boundary of $\mathcal{P}$. If the count is odd for all four lines, then the cell lies inside the polygon. Further, we sample random grids from the grid graph and traverse the neighbour grids in all four directions to extract the rectangle with the largest area possible. This results in a set of disjoint rectangles $d_1, \ldots, d_q$, such that the union of $d_i$'s forms the polygon $\mathcal{P}$, see Fig.~\ref{fig:decomposition}(a). 

\noindent\paragraph{Computing Space-filling curves} We utilize the extension~\cite{gilbert} of~\cite{zhang2006} to obtain the generalized rectangular space-filling curves for rectangles that has a side of length which is not a power of two. In this approach, a rectangle (with horizontal length: $a$ and vertical length: $b$) is spilt into three regions `up', `right', and `down', recursively until a trivial path is produced; see Fig.~\ref{fig:decomposition}(b).

\begin{figure}\centering
  \includegraphics[scale=0.45]{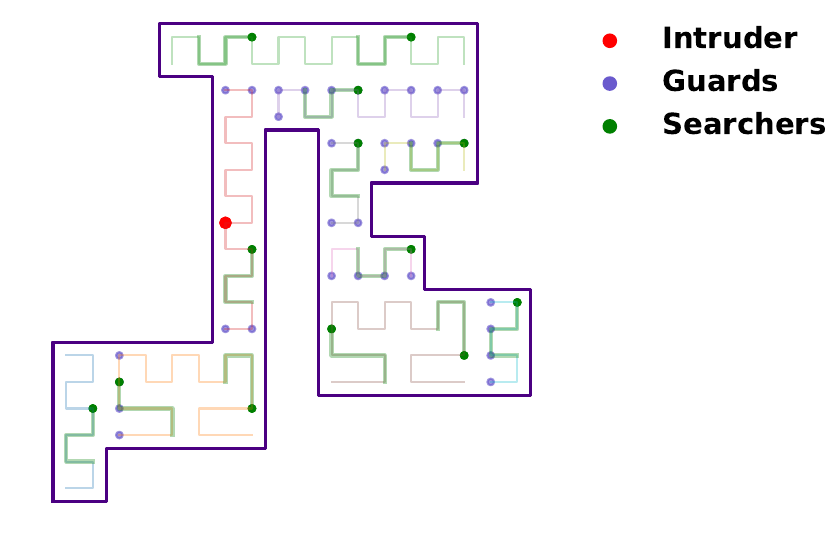}
  \caption{Illustration for the arrangement and distribution of search robots in Space-filling curve approach. The polygonal region is decomposed into many sub-rectangular regions ($d_i$) and each sub-rectangular region is filled with a space-filling curve $c_i$. 
  The trail behind each agent shows the trajectory during its past few time steps.}  \label{fig:sfc}
\end{figure}

Let the sensing range of the robot be $l^s/2$. For each rectangular region $d_i$, we obtain the grid graph $G_i = G(s_i)$ using smaller square grids $g_i$ of dimensions double the size of the robot's sensing range $l^s$. Then, we obtain the dual graph $\mathcal{D}_{G_i}$ of $G_i$, where the dual graph is obtained by connecting the centroids (nodes of $\mathcal{D}_{G_i}$) of each $g_i$ to its neighbouring grids' centroids through an edge (edge of $\mathcal{D}_{G_i}$).
Using the nodes of $\mathcal{D}_{G_i}$, we obtain the modified Hilbert's rectangular space-filling curves \cite{gilbert} $c_i$, for each sub-region $d_i$. When two sub-regions $d_i$ and $d_j$ are adjacent to each other, they share a common edge. We define these common edges as junctions $J_{ij}$, as shown by dashed black lines in Fig.~\ref{fig:illustration}(b). The curve $\mathcal{C}$ ensures complete spanning of all the regions' $d_i$ in $\mathcal{P}$. However, each $c_i$ is disconnected at $J_{ij}$ (see Fig.~\ref{fig:sfc}). The robots utilize these disjoint curves to move along it and search for the intruder while staying inside the same $d_i$ for the entire time.

\noindent\paragraph{Assigning robots and search process}  The robots are divided into two groups: search robots and guard robots of size $k_s$ (green dots in Fig.~\ref{fig:sfc}) and $k_g$ (purple dots in Fig.~\ref{fig:sfc}), respectively. When $k_g=0$, we define this case as a space-filling curve approach (SFC) whereas if $k_g\neq 0$, then we define it as a Guarded space-filling curve approach (SFC-G). The number of robots assigned to each rectangular area, $d_i$, is equal to the proportion of the area of the sub-rectangular with respect to the total area of the polygon. The search robots keep patrolling the assigned path, $c_i$. However, the guard robots remain stationary at the junction $J_{ij}$ to prevent re-contamination. For a detailed illustration of the arrangement, see Fig.~\ref{fig:sfc}.

Consider the grid graph $G(\mathcal{P}) = \bigcup G(d_p)$, that contains a total of $n\times m$ nodes (line $1$ in Alg.~\ref{alg:sfc}). In order to determine the time required to search the static/moving intruder using $k$ robots, where $k = |\mathcal{R'}|$, we determine the number of steps that the robots are required to move to find the intruder. 
We start with a total number of robots that is equal to the total number of grids $(nm)$ in the polygon and continue until $k$ becomes equal to the sum of the number of $d_i$ regions and the number of guarding robots, i.e. $k = |\{d_i|d_i\epsilon\mathcal{P}\}| + k_g$. The simulation finishes if the searcher and the intruder share the same cell. The pseudo-code in Alg.~\ref{alg:sfc} (line 5-8) shows the logical execution of each robot's task.

\begin{algorithm}
    \caption{Space filling approach}
    \begin{algorithmic}[1]
        \State $grids \gets n\times m\;nodes $
            \State $search\_state \gets False $
            \While {\textbf{not} $search\_state$}
                \For {r \textbf{in} robots}
                    \If {`r' location = intruder's location}
                    \State $search\_state \gets True$
                    \EndIf
                    \State Move the robot `r' along SFC
                \EndFor
                \If {$search\_state$}
                    \State store search number with the time taken
                \EndIf
            \EndWhile             
    \end{algorithmic}
    \label{alg:sfc}
\end{algorithm}

\noindent \emph{Runtime:} If there are $k$ robots. The algorithm for space-filling curves approach has a worst-case time complexity $\mathcal{O}(k)$.

\subsection{Random search approach (RS)}\label{sec:random}

The space-filling approach is partially deterministic in nature. If either end of $c_i$ is revealed to the intruder, it may successfully avoid the searchers. Therefore, we propose a randomized approach, the random search algorithm (RS). Here, $k$ robots are distributed randomly uniformly in the region $\mathcal{P}$. Each robot $r_i$ randomly uniformly samples one target grid cell (say, $g_t$) inside $\mathcal{P}$, and plans a path ($x_i$) using A$^\star$ algorithm \cite{Peter1968} connecting its current position to $g_t$. 

The random search approach has an added cost due to the possibility of repeated visits to previously explored cells. For instance, when two larger spaces are connected through a narrow channel, robots will search the narrow region multiple times more often. To avoid this, we redesign the heuristic function of the A$^\star$ algorithm by accommodating the retraversal cost during the path planning. Whenever a robot visits a grid, it raises its cost ($L_{g_i}$) by $0.05$. So, any robot(s) that plans a path in the future will avoid travelling through the previously visited grids, reducing the frequency of revisits. Such paths are obtained by minimizing the path length to the target grid and the cost (the total sum, $\Sigma L_{g_i}$) of revisiting the grids along the path. If a robot reaches its target grid, irrespective of other robots' status, it re-samples a new location to continue searching for the intruder. 

\begin{algorithm}
    \caption{RS and CRS}
    \begin{algorithmic}[1]
        \State $search\_state \gets False $,  $grids \gets n\times m\;nodes $
        \State CRS $\gets$ True 
        \State $costs \gets$ array of cost of traversing through a grid
        \While {\textbf{not} $search\_state$}
            \State Update costs of each grid
            \For {r \textbf{in} robots}
                \If {r reached its destination}
                    \State Sample new location uniformly randomly.
                    \State Plan path
                \EndIf
                \If {all robots at destination and CRS == True}
                \State Allocate new targets via Hungarian algorithm   
                \State Plan path
                \EndIf
                \State Update 'r' location along its path
            \EndFor
            \If {intruder is found}
                \State $search\_state \gets True $
            \EndIf
        \EndWhile             
    \end{algorithmic}
    \label{alg:rs}
\end{algorithm}

In this approach, since the searchers are not constrained to move inside sub-regions of $\mathcal{P}$, guarding robots are not required. Thus reducing the extensive need for robots, and hence, the algorithm works for any number of robots. The pseudo-code in Alg.~\ref{alg:rs} shows the logical execution of the random search algorithm.

\noindent \emph{Runtime:} Since in the worst case, each robot can have a path length equal to the total number of grids in the grid graph. The A$^\star$ itself will have a time complexity of $\mathcal{O}(nm)$. Therefore, if there are $k$ robots, this approach has a worst-case time complexity of $\mathcal{O}(knm)$.
\begin{figure}
  \centering
  \includegraphics[scale=0.5]{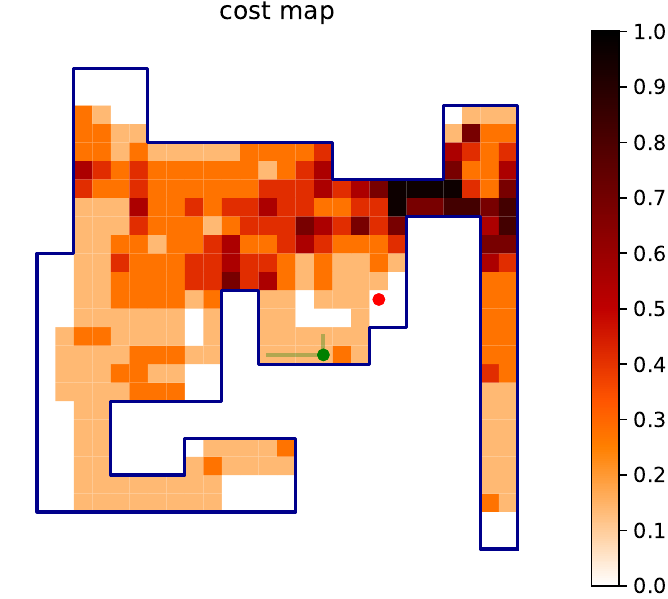}
  
  \caption{A snapshot of cost map from the random search algorithm. The intruder and the robot are depicted in red and green circles, respectively. The trajectory of the searcher is shown with a green trail and the intruder with a red trail. The grid colours indicate the cost of re-traversal. The higher (brighter) the intensity, the lower the cost.}
  \label{fig:rs}
  
\end{figure}

\subsection{Cooperative random search approach (CRS)}\label{sec:crs}

In this approach, the search process is the same as RS however, the robots undertake target-matching once all the robots reach their respective destinations. The robots use the Hungarian task matching algorithm \cite{kuhn1955} to assign a target among from the newly generated set of target locations. The assignment reduces the total path length of the agent paths.  

Say, there are $k$ robots that sample uniformly randomly $k$ grid cells. If any robot discovers the intruder along its path then the algorithm terminates. Otherwise, after all the robots reach their random targets, they sample the next $k$ grid cells uniformly randomly. In Alg.~\ref{alg:rs}, see the lines 11--14 that include the cooperative aspect to the RS method.

\noindent \emph{Runtime:} 
If there are $k$ robots, the Hungarian algorithm has a worst-case time complexity of $\mathcal{O}(k^3)$. However, to minimize the path length, robots first plan paths using A$^\star$ which has a worst-case time complexity of $\mathcal{O}(k^2nm)$. Therefore, the worst-case time complexity of this approach becomes $\mathcal{O}(k^3+k^2nm)$. Since we assume that $k\ll nm$, the worst-case time complexity for the cooperative random search approach is $\mathcal{O}(k^2nm)$.

\section{Simulation Results}\label{sec:experiments}

In this Section, we evaluate the performance of the proposed methods through Monte-Carlo simulations.  We analyze the change in the intruder discovery time by increasing the number of searchers and changing the geometric property by regulating (a) the number of spikes, $\beta$, (b) the area and (c) the shape of the polygon $\mathcal{P}$. We assume that the search robots have a limited sensing range and do not have apriori information (position) about the intruder. Also, we model the intruder to move randomly in the domain, $\mathcal{P}$. The codes for all three methods can be found at~\cite{suppMaterial}.

\subsection{Simulation setup}

\begin{figure}
  \centering
  \includegraphics[width=0.8\linewidth]{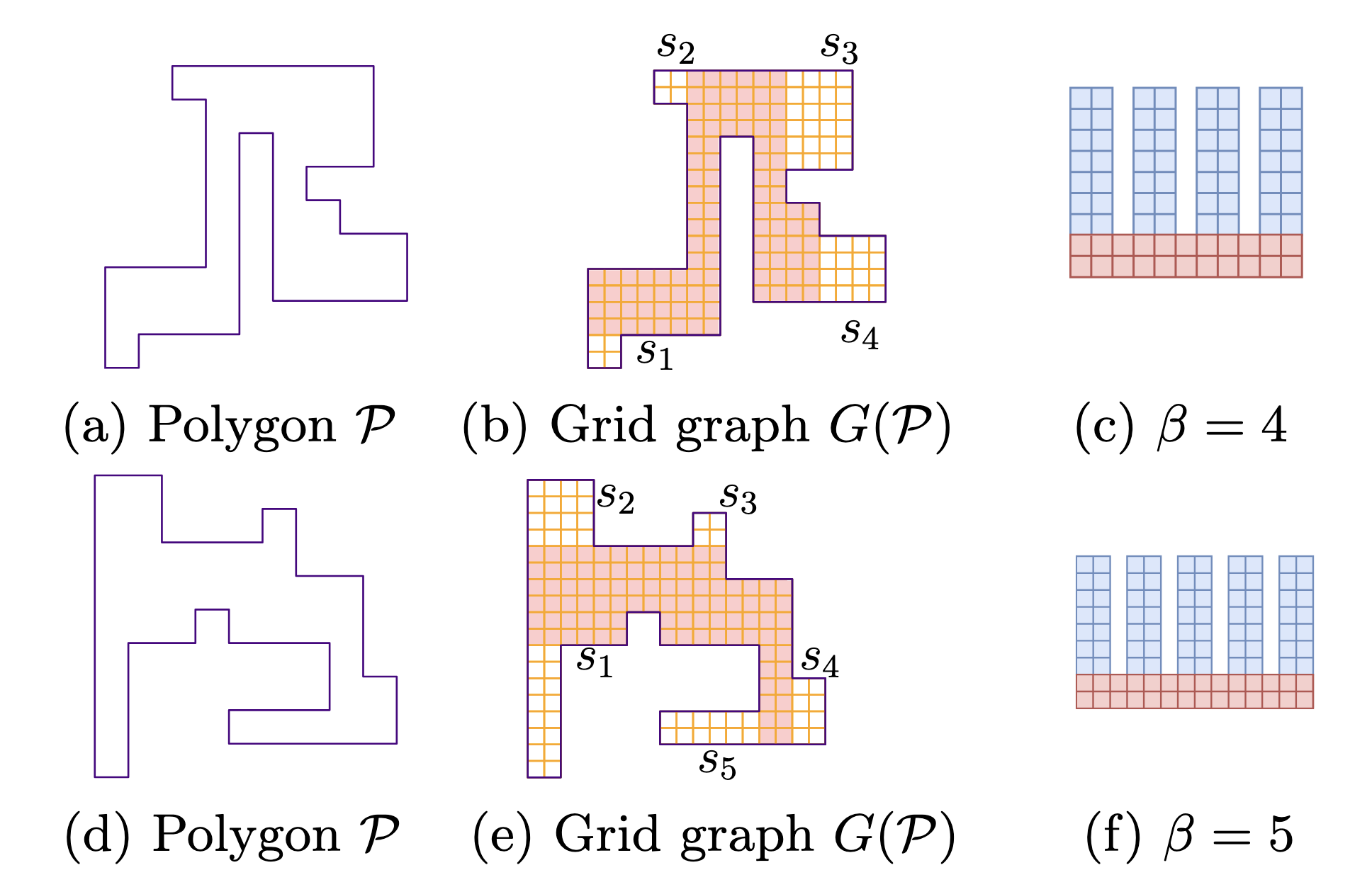}
  \caption{The region $\mathcal{P}$ with $\beta = 4$ spikes shape (top row), and $\beta = 5$ spikes (bottom row). Both the polygons have a minimum edge length of $l^b = 10m$. The orange grid shows the grid graph, $G(\mathcal{P})$, with grid cells of the size, $l^s = 5m$. The light red shaded region is the base of the comb structure both in the original polygon and the schematic illustration beside the corresponding polygons. The un-shaded regions in the polygons are the spikes $s_i$.}
  \label{fig:spike}
\end{figure}

The comb structure defined above in Sec.~\ref{sec:hardness} spans the whole space of simply connected orthogonal polygons via the number of spikes, $\beta$. Therefore, we model the environment's complexity in terms of $\beta$; the higher $\beta$, the more complex the environment. The orthogonal polygons of equal area are obtained through the polynomial time Inflate-Cut algorithm~\cite{APTomas}, as shown in Fig.~\ref{fig:spike}(a), \& (d). The polygon $\mathcal{P}$
has minimum edge length varied from $l^b=5m$ to $l^b=30m$ across various experiments, and each robot has a fixed sensing range of $l^s/2 = 2.5m$. We discretize the domain into cells. Each cell is of $l^s=5m$ length.
The robots move from the centroid of one cell to another in a unit time step. To our knowledge, no known methods exist to determine or regulate the number of spikes in a given simple orthogonal polygon. Therefore, we limit our exploration to manually counting the number of spikes for any random simple orthogonal polygon through visual inspection.

\noindent\paragraph{Counting spikes, $\beta$} We classify a part of the polygon as a spike only if two adjacent edges of a spike form a subset of the polygon's boundary edge set and any of the third edge of the spike must be a fractional part or complete edge of the next adjacent edge in the boundary edge set. Whereas the fourth side of the spike must be open (see Fig.~\ref{fig:spike}).

The performance of the proposed heuristics is validated on two types of scenarios -  static intruder  (S-I)  and moving intruder (M-I). The performance of the algorithms can vary based on the number of search robots, the number of spikes $\beta$, and the shape of the polygon with an equal number of spikes. Therefore, we perform Monte-Carlo simulations (100 simulations) for each case while varying the number of searching robots, $k$, the number of spikes $\beta$, and the shape of the polygon for each scenario (S-I and M-I) to compare the average performance of each approach. 

\subsection{Baseline}

We compare the performance of the proposed algorithms with a baseline approach where the searchers know the position of the intruder at every instant of time, and they compute the shortest path (Dijkstra algorithm~\cite{Dijkstra1959}) to intercept the intruder. This forms the lower-bound solution. Ideally, we would like the heuristic solution to be as close as possible to the optimal solution (baseline solution). 

\subsection{Effect of increase in the number of search robots}
Consider the scenario as shown in Figure~\ref{fig:spike}(a), where the polygon $\mathcal{P}$ has $\beta = 4$ spikes depicted with uncoloured grids in Fig.~\ref{fig:spike}(b). The polygon has a total area of $152$ cells. Given $4$ spikes, we vary the number of robots from $2$ (or minimum possible for each approach) to $152$ (or maximum possible corresponding to the number of cells) as $2,5,8,\ldots,60$ with an interval of three robots. For a given number of robots, we iterate 100 simulations with randomly generated positions of the searchers and the intruders. For each simulation, we evaluate the performance of the four methods -- SFC, SFC-G, RS and CRS with respect to the baseline strategy.

\begin{figure}
    \centering
    \includegraphics[width=0.45\linewidth]{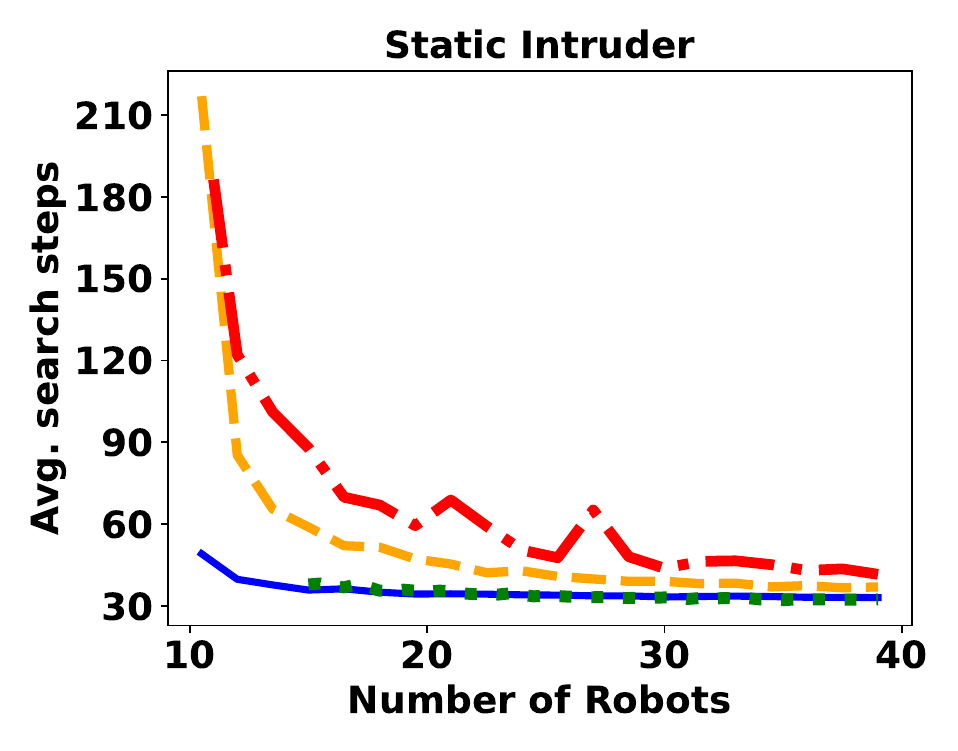}
    \includegraphics[width=0.45\linewidth]{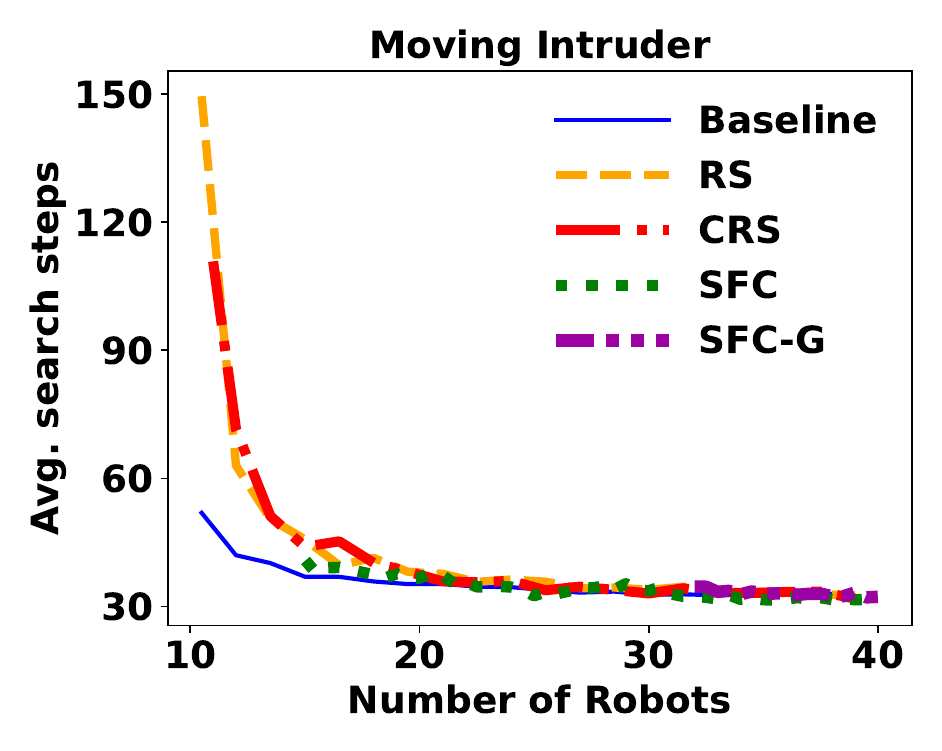}

    \caption{Effect of increase in the number of searchers on search steps for different search strategies on 4-spiked polygon. (a) for static intruder case and (b) for moving intruder case.}
    \label{fig:line_5spike}
\end{figure}

Figure~\ref{fig:line_5spike} shows the average number of search steps required to intercept the intruder for the different number of robotic searchers under S-I and M-I scenarios. From the figure, we can see that in the case of RS, CRS and Baseline strategies, solutions exist for the minimal number of searchers (2). However, for the SFC strategy, for S-I case, a minimum of 10 searchers are required to determine a solution and 44 searchers to determine a solution in the case of M-I problem. The number of searchers required in the case of SFC is high because of the rectangulation of the polygonal area as shown in Fig.~\ref{fig:decomposition}(a). For each rectangle, we need at least one robot. While in the case of SFC-G, guard robots are also deployed for each junction between the rectangles. This further increases the number of robots required. 

\begin{figure}
    \centering
    \includegraphics[scale=0.4]{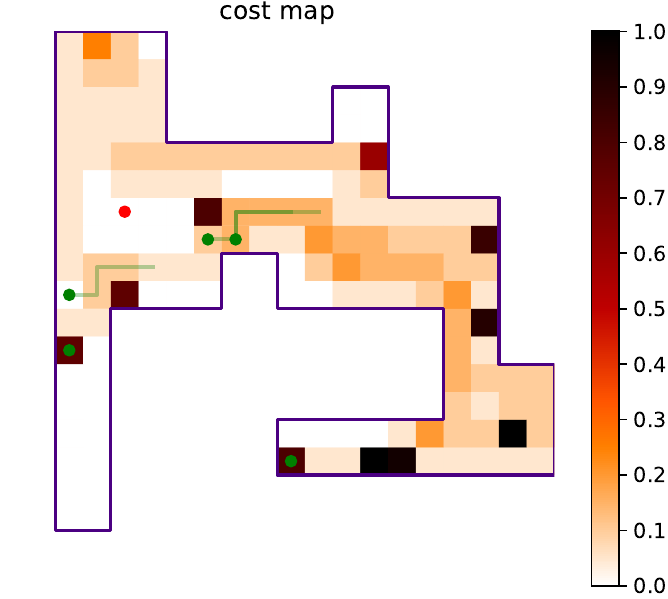}
    \caption{Snapshot from cooperative random search for a static intruder. Some grids appear dark due to the accumulation of revisiting costs during the waiting time of search robots until all search robots reach their goal.}
    \label{fig:crs_waiting}
    
\end{figure}

A natural hypothesis between RS and CRS is that the CRS will perform better than RS. However, the results show a different trend, where RS is performing better than CRS in the S-I case, while CRS is performing better than RS in the M-I case. For S-I, this happens because, in the case of RS, the robots are randomly given goals and upon reaching the goal they are immediately given another random goal. However, in the case of CRS, the robot that has reached a goal early has to wait for all the other robots to reach their goal and then generate the next set of goals. Due to this, there is a significant loss of search time steps for the agents as shown in Fig.~\ref{fig:crs_waiting}. This loss is reflected in the results in terms of increased search time. In the M-I case, the motion of the intruder helps the searchers to capture it quickly, given that the number of searchers is much higher. When the number of searchers is high, in the CRS approach, the robots have to wait for longer at their target grids for other searchers to reach soon. However, by the time all robots reach their target grids, the randomly moving intruder gets discovered by the distributed waiting searchers.

When the minimum number of robots available is equal to the number of rectangles in the arena, then SFC performs far better than RS, CRS and SFC-G. The results are very close to the baseline strategy. This is because the robots need to search fewer number of cells. Note that, in the case of static intruder case, we do not include the SFC-G method because guards are not required in this case.

\subsection{Effect of change in the shape of the polygon}

\begin{figure}
    \centering
        \includegraphics[width=0.45\linewidth]{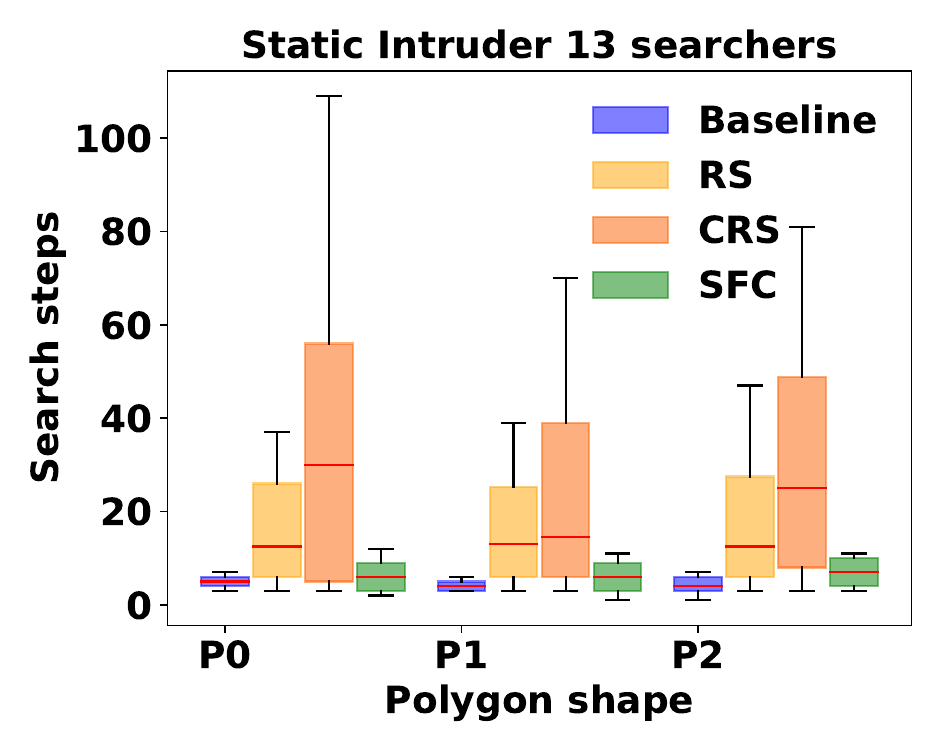}
        \includegraphics[width=0.45\linewidth]{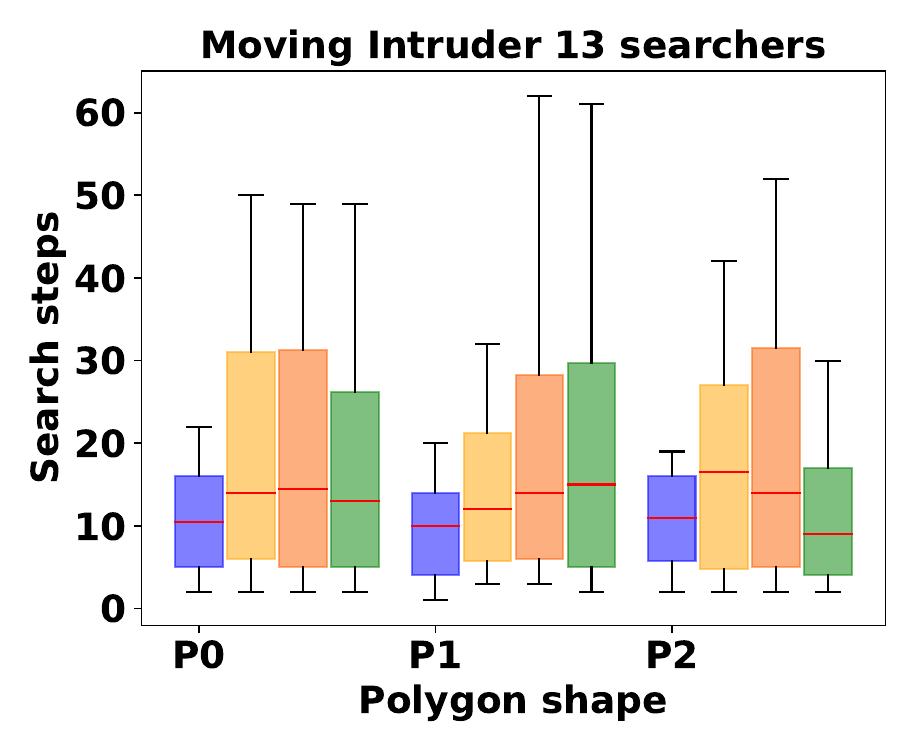}
        \includegraphics[scale=0.2]{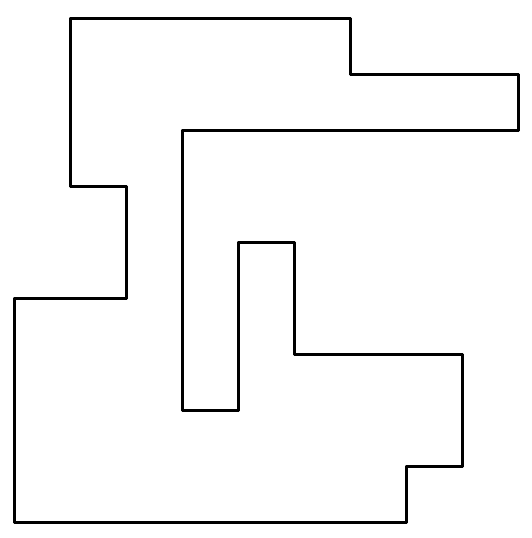}
        \put(-35,-10){P0}
        \includegraphics[scale=0.2]{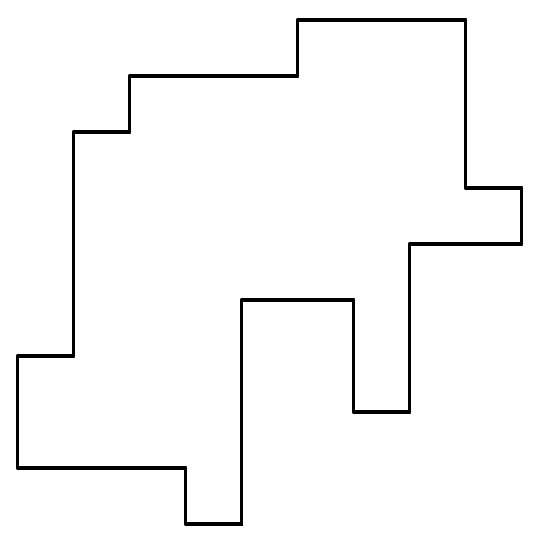}
        \put(-35,-10){P1}
        \includegraphics[scale=0.2]{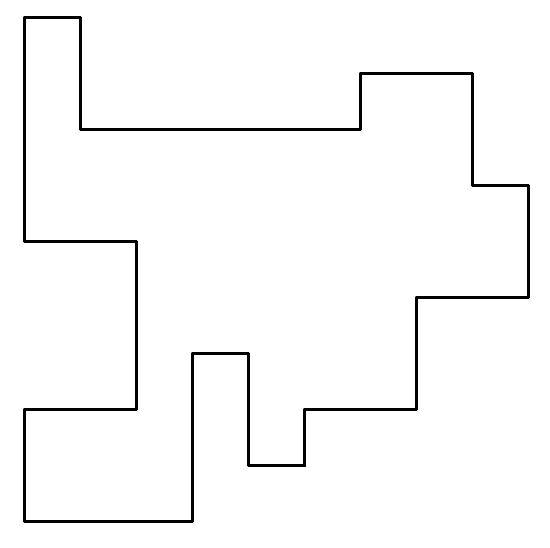}
        \put(-35,-10){P2}
    \caption{Effect of change in the shape of the polygon (area $4400\;unit^2$ and $\beta=5$) on search steps for different search strategies and $k=13$ search robots, (a) for static intruder case and (b) for moving intruder case.}
    \label{fig:changing_shape}
\end{figure}

Further, changing the shape of the polygonal region is similar to re-arranging the spikes, and it should not affect the search time required by the $k$ robots. To validate the same, we consider $13$ search robots in three polygons of varying shapes and equal area ($4400\;unit^2$) that fall in the category of $\beta=5$. As shown in Fig.~\ref{fig:changing_shape}, for each proposed approach, the search time remains statistically indifferent with changing shapes, irrespective of whether the intruder is static or moving. In the case of a static intruder, regardless of the shape of the polygon, the search time closest to the Baseline approach is in the case of SFC. Whereas, RS performs better than CRS and worse than SFC. Moreover, the intruder searching is faster in the case of the Baseline and SFC approach when the intruder is static. The RS performs indifferently irrespective of the intruder being static or moving. However, CRS is faster in searching for the intruder when it is moving. Hence, we conclude that the shape of the polygon does not affect the performance of the approaches proposed in this work.
\subsection{Effect of increase in area of the polygon}

\begin{figure}
    \centering
        \includegraphics[width=0.45\linewidth]{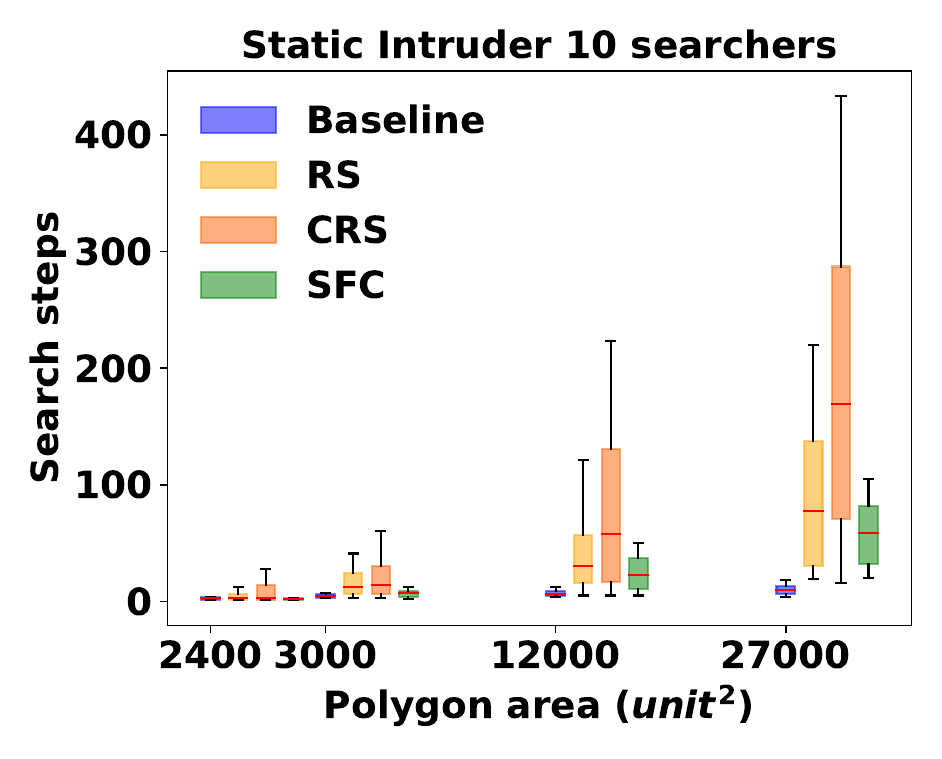}
        \includegraphics[width=0.45\linewidth]{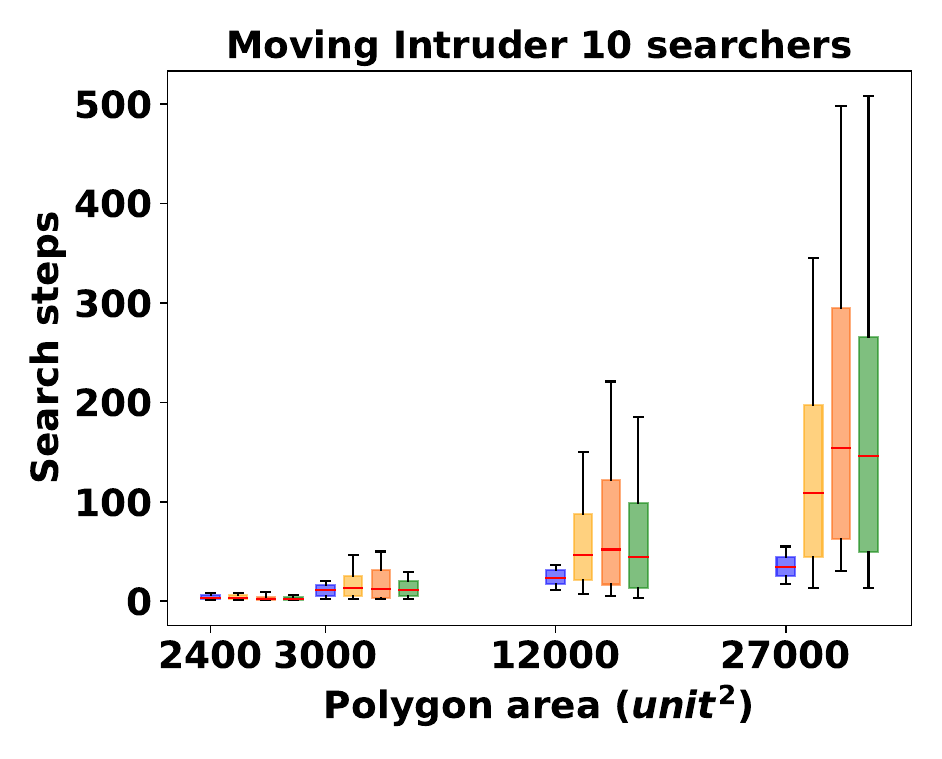}

    \caption{Effect of change in the area of the polygon (fixed shape and $\beta=5$) on search steps for different search strategies and $k=10$ search robots, (a) for static intruder case and (b) for moving intruder case.}
    \label{fig:changing_area}
\end{figure}
Next, we consider a polygon with a fixed shape in the category of $\beta=5$. When there were $10$ searchers, and the area of the search region was increased, the search time increased irrespective of the intruder being static or moving in all the different approaches as shown in Fig.~\ref{fig:changing_area}. However, the performance of the different algorithms remained similar, as in the case of changing shapes of the polygon with a static intruder in it. In contrast, with the increasing area of the region, all the approaches took longer to search for the moving intruder than for the static intruder.

\subsection{Effect of increase in the number of spikes}

\begin{figure}
    \centering
        \includegraphics[width=0.45\linewidth]{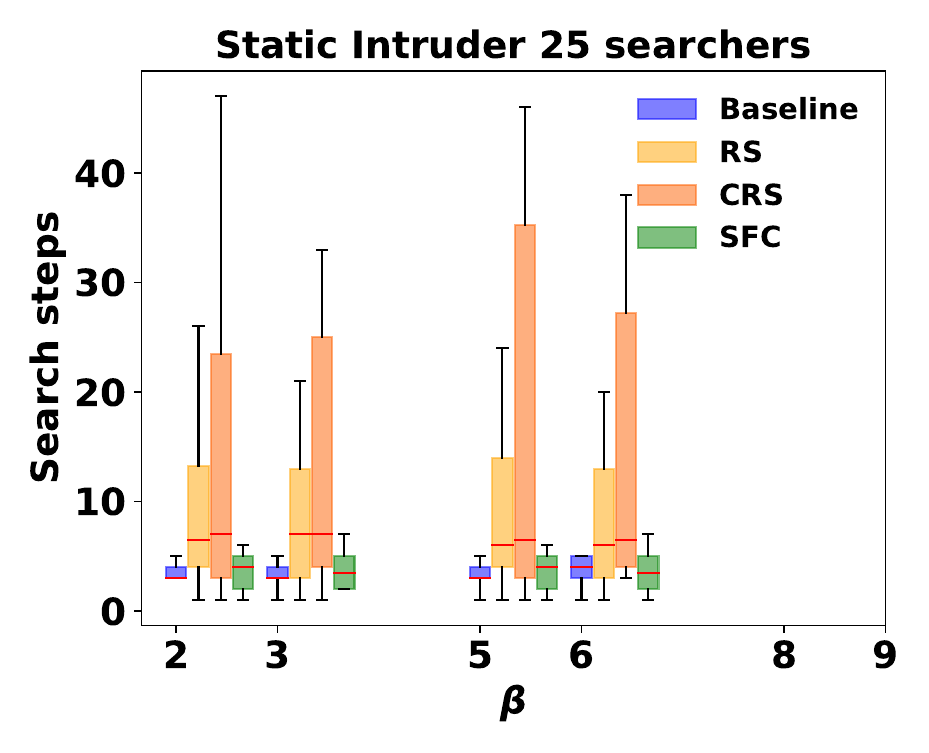}
        \includegraphics[width=0.45\linewidth]{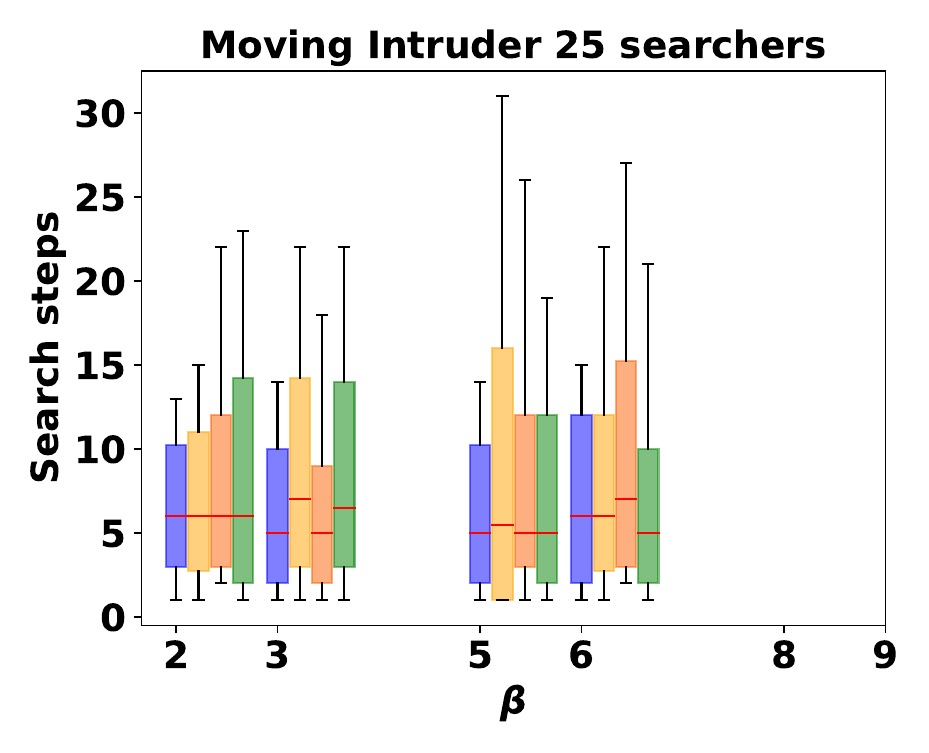}
    \caption{Effect of increase in the number of spikes on search steps for different search strategies and $k=25$ search robots, (a) for static intruder case and (b) for moving intruder case when there was $160$ total number of cells.}
    \label{fig:bar_spike}
\end{figure}

The geometric complexity of the search region is captured by the number of spikes. To understand the effect of geometric complexities on search time in the given scenario, we increase the number of spikes from $2$ to $6$ and see the change in performance as shown in Figure~\ref{fig:bar_spike}. The total area of the polygon is fixed to $4000\;unit^2$; however, the shape of the polygon changes with the changing $\beta$. We can observe that when there were $25$ search robots, both for static and moving intruders, the number of search steps required to accomplish the task remains unaffected as the number of spikes, $\beta$, increase from $2$ to $6$ in all four approaches. This simple illustration shows that with an increase in geometric complexity, the search time is unaffected. However, we believe that increased search overlap percentage (resulting from increasing $\beta$) should significantly reduce the effective search in new regions and hence increase the intruder discovery time. To validate our hypothesis, more work is required to generate polygons with desired $\beta$, which is beyond the scope of this work. Furthermore, we also observe that the searching of a moving intruder becomes faster with the RS and CRS approach as compared to the static intruder. Meanwhile, searching becomes slower for the moving intruder with the Baseline and SFC approach.

\section{Conclusions and Discussions}\label{sec:conclusion}

In this paper, we study the MRS and DMRS problems inside a simply connected orthogonal polygon. We have shown that MRS is \textsf{NP}-hard, and conjectured that DMRS is also \textsf{NP}-hard. On the positive side, we proposed three algorithmic methods to provide solutions to the problem. We performed Monte-Carlo simulations and evaluated the performance of these methods and also compared them with a baseline. The RS and CRS methods are effective even with a minimal number of search robots; however, they need longer search time. The space-filling methods provide solutions close to the baseline. However, it requires a minimum number of robots equal to the decomposed number of rectangular regions in $\mathcal{P}$. With an increased number of robots, the performance is almost near the baseline. In most scenarios, finding a moving intruder via RS and CRS approaches is quicker.
In contrast, finding a static intruder via Baseline and SFC approaches is quicker. Also, we conclude that a few geometric properties like $\beta$ and area can severely affect the searching, whereas in other cases, for instance, the shapes with the same geometric properties do not affect the searching significantly.

The proposed approach can be extended further to study the effect of multiple intruder cases. Also, the problem can be extended to an intelligent intruder case, where the intruder wants to evade the searcher robots. Similarly, the study can be extended to more complex search spaces, such as polygon with holes. From the simulations, it can be seen that SFC is quite effective, but, the minimum number of robots required depends on the rectangular decomposition. Therefore, another interesting study can be to develop algorithms for minimizing the number of rectangles constructed in $\mathcal{P}$. Ideally, it is possible to bypass the rectangulation by using space-filling curves for more complex regions, which leads to a rather complex mathematical problem.

\section*{Acknowledgments}
This work has been supported by IISER Bhopal and IIT Bombay.

\bibliographystyle{splncs04}
\bibliography{references}






\end{document}